\documentclass[10pt,twocolumn,letterpaper]{article}

\usepackage{iccv}
\usepackage{times}
\usepackage{epsfig}
\usepackage{graphicx}
\usepackage{amsmath}
\usepackage{amssymb}

\DeclareMathOperator*{\argmax}{arg\,max}

\DeclareMathOperator*{\degree}{^\circ}
\usepackage{xcolor, soul}
\sethlcolor{yellow}

\usepackage[pagebackref=true,breaklinks=true,letterpaper=true,colorlinks,bookmarks=false]{hyperref}

\iccvfinalcopy 

\def\iccvPaperID{25} 

\begin{document}


\author{Max Lennon\thanks{equal contribution}, Nathan Drenkow\footnotemark[1], Philippe Burlina \\
The Johns Hopkins University Applied Physics Laboratory\\
11100 Johns Hopkins Road, Laurel, Maryland 20723\\
{\tt\small Nathan.Drenkow@jhuapl.edu}
\\
}


\title{Patch Attack Invariance: How Sensitive are Patch Attacks to 3D Pose?} 
\maketitle

\begin{abstract}
\vspace{-.4cm}
Perturbation-based attacks, while not physically realizable, have been the main emphasis of adversarial machine learning (ML) research. Patch-based attacks by contrast are physically realizable, yet most work has focused on 2D domain with recent forays into 3D. Characterizing the robustness properties of patch attacks and their invariance to 3D pose is important, yet not fully elucidated, and is the focus of this paper. To this end, several  contributions are made here: A) we develop a new metric called \emph{mean Attack Success over Transformations (mAST)} to evaluate patch attack robustness and invariance; and B), we systematically assess robustness of patch attacks to 3D position and orientation for various conditions; in particular, we conduct a sensitivity analysis which provides important qualitative insights into attack effectiveness as a function of the 3D pose of a patch relative to the camera (rotation, translation) and sets forth some properties for patch attack 3D invariance; and C), we draw novel qualitative conclusions including: 1) we demonstrate that for some 3D transformations, namely rotation and loom, increasing the training distribution support yields an increase in patch success over the full range at test time. 2) We provide new insights into the existence of a fundamental cutoff limit in patch attack effectiveness that depends on the extent of out-of-plane rotation angles. These findings should collectively guide future design of 3D patch attacks and defenses.

\end{abstract}

\section{Introduction}

Over the past several years, deep learning (DL)~\cite{lecun2015deep} methods have enabled continuous success at tackling problems including classification~\cite{krizhevsky2017imagenet,he2016deep}, detection~\cite{Ren2015FasterRT,Redmon2015YouOL}, segmentation~\cite{pekala2019deep,ronneberger2015u}, medical diagnostics~\cite{burlina2019automated,burlina2018utility}, and game playing~\cite{mnih2015human,silver2017mastering}, while achieving performance on par with humans and vastly exceeding legacy machine vision systems~\cite{vyas2013estimating,burlina1995context}. Current DL research has now  turned to more challenging issues including trusted AI and more difficult cases including privacy attacks~\cite{shokri2017membership}, open set recognition~\cite{scheirer2012toward,geng2020recent}, achieving low shot learning~\cite{ravi2016optimization,burlina2020low},  AI fairness~\cite{burlina2020addressing}, and importantly, and also the focus of this work,   adversarial machine learning  (AML)~\cite{goodfellow2014explaining,carlini2017adversarial}.

{\bf Prior work:} AML has  motivated intense work  in the past years focused on methods addressing both attacks and defenses, in what has become a perpetual cat-and-mouse game. The overarching majority of  approaches to AML directed at machine perception have focused on attacks that affect entire images,  are optimized to work on specific images/scenes, and use additive imperceptible perturbations (e.g.~\cite{szegedy2013intriguing},~\cite{goodfellow2014explaining},~\cite{moosavi2016deepfool},~\cite{moosavi2017universal}). While these  remain  the main focus of adversarial machine learning research in the area of visual perception, perturbation-based attacks assume that the adversary has electronic access, and are not otherwise physically realizable. By contrast,  patch-based attacks are realizable in the physical domain and are instead based on attacks that are locally-confined to a specific part of the image and designed to affect a wide set of images with specific classes of objects. These include both patch and occlusion based attacks~\cite{brown2017adversarial}. 
We focus here on these attacks since they are implementable in the physical domain via simple printing and placement of optimized patches or via occlusions, easily introduced  in a scene. These patch attacks are of interest for many areas of autonomy, being of significant concern for applications in robotics/vehicular autonomy\cite{wu2020physical}, remote sensing\cite{czaja2018adversarial}, and other AI applications\cite{shah2018susceptibility}.

\begin{figure}[t!]
\begin{center}
    \begin{tabular}{c}
        \includegraphics[width=.9\linewidth]{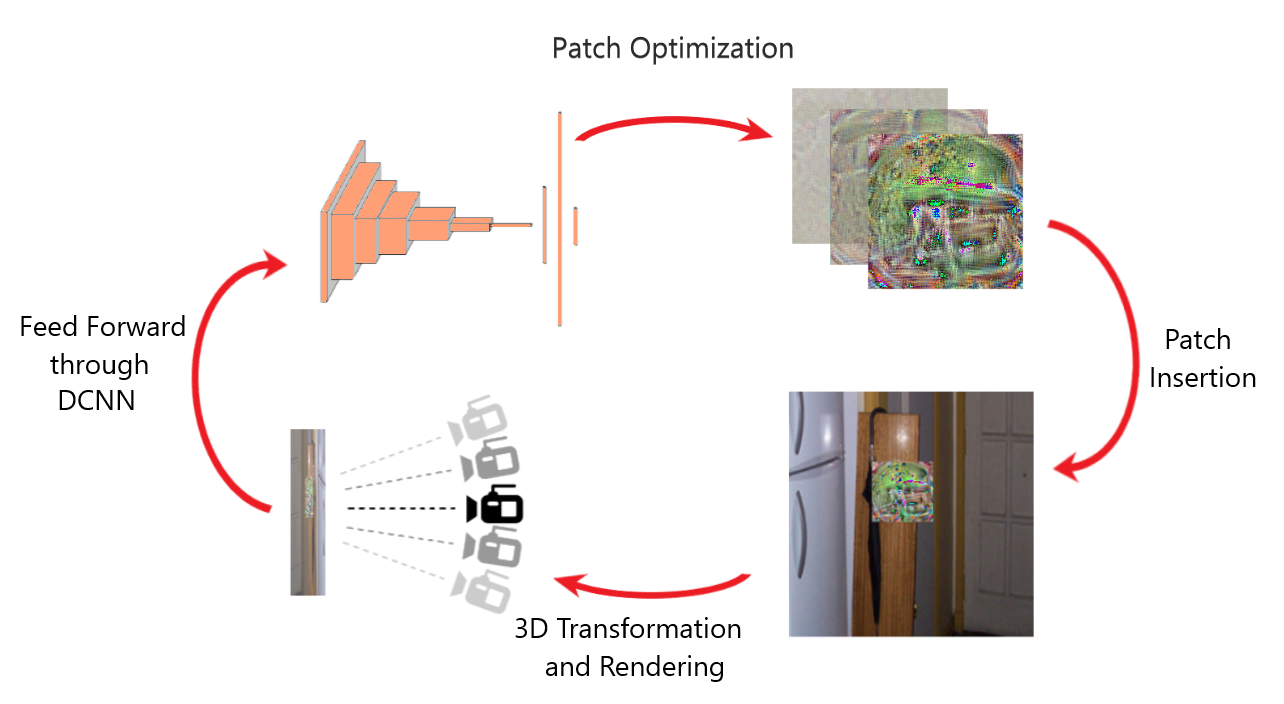}
    \end{tabular}
\end{center}
  \caption{3D patch optimization.}
\label{fig:constrained_eot}
\end{figure}

Examples of  work focusing on such attacks includes~\cite{athalye2017synthesizing,brown2017adversarial} which implemented a loss expressed as an expectation taken over a set of possible geometric transformations including 2D rotations, translation, and scale~\cite{athalye2017synthesizing}.  Patch attack development originally focused on fooling recognition systems~\cite{eykholt2018robust, kurakin2016adversarial, sharif2016accessorize} but have since been extended to various forms of object detection~\cite{thys2019fooling, zolfi2021translucent, wu2020dpattack, zhao2020object, wu2020making, chen2018shapeshifter}. Related work includes the study of physical adversarial attacks for vehicle detectors using a simulator (Carla)\cite{wu2020physical}. Prior work here lacks a detailed 3D sensitivity analysis of realizable patch-based attacks. The objective of our study is to investigate robustness characteristics of patch attacks in a 3D setting and specifically with regard to pose parameters, a problem which is still in need of deeper investigation.

{\bf Contributions:} Our novel contributions pertain to physically realizable attacks that are specifically focused on adversarial patches and are optimized for 3D pose.  We characterize the sensitivity of the attack effectiveness as a function of the pose of the patch (rotation, translation) with respect to the camera frame. We define a new robustness metric (called \emph{mean Attack Success over Transformation, mAST}) based on mean (taken over classes) average robustness (taken over the kinematic transformation span). We further derive important qualitative characterization to inform future research in improving attacks and defenses.

\begin{figure*}[h]
    \centering
    \includegraphics[scale=0.4]{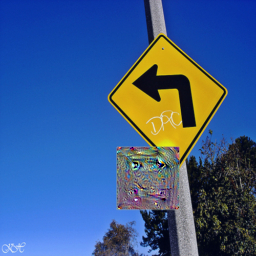}
    \includegraphics[scale=0.4]{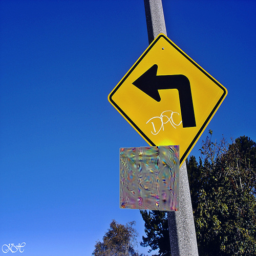}
    \includegraphics[scale=0.4]{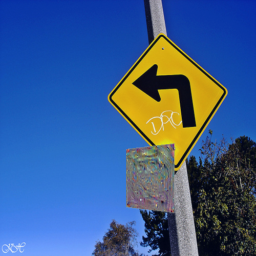}
    \includegraphics[scale=0.4]{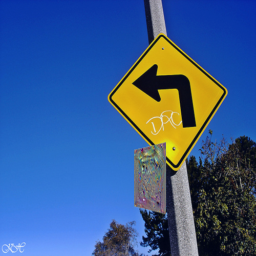}
    \caption{Patches targeting the "mailbox" class inserted into a base image at (left to right) $ 0\degree,\ 20\degree,\ 40\degree, \text{ and } 60\degree\ $ of yaw.}
    \vspace{-4mm}
\end{figure*}

\section{Methods}

We first provide our approach for performing patch optimization over varying camera viewpoints which might occur as a result of, e.g., platform motion. Our general strategy for targeted patch attack generation is to find a patch that is optimal in expectation over both images and transformations, i.e., follows a pattern of optimization as follows:

\begin{equation}
\label{eq:general}
    \argmax _{q} \mathbb{E}_{I\sim{{\cal I}}, T\sim\mathcal{T}}~ \Phi(y_t|\mathcal{A}_T(q, I))
\end{equation}

\noindent where the terms denote: the patch $q$, image $I$, transformation $T$, application method $\mathcal{A}_T(\cdot)$, targeted label $y_t$, predictive model $\Phi(\cdot)$ (such as a deep convolutional neural network). 

Consider the situation of co-planar patches.
For this, assume that a camera images a patch, where the patch is considered fixed and the camera has undergone 3D rigid motion (i.e., a transformation from SE(3) the special Euclidean group of 3D rotations and translations) from time $0$ to time $t$. The coordinates of a point lying on a patch at time t, denoted by $P_t$, where $P_t = [X, Y, Z]^T$ and $P_t$ is expressed in the camera frame of reference. While undergoing motion, the point coordinates satisfy:
\begin{equation}
    P_t = {\bf R} P_0 + {\bf T}
\end{equation}
with the rotation ${\bf R}$ and translation ${\bf T}$ aligning the patch coordinate frame of reference and the image frame of reference, from time $0$ to time $t$; $P_0$ denotes the coordinate at time 0.

We also denote by 
\begin{equation}
 p_t = c(P) 
 = 
 \begin{bmatrix}
 f_x \frac{X_t}{Z_t} \\
 f_y \frac{Y_t}{Z_t}   
 \end{bmatrix}
 \label{eq:point_proj}
\end{equation}

\noindent
the coordinates of the 2D perspective projection of the 3D point $P$  into the camera (with $f$ denoting camera focal length). Then it is well-known that $p_0$ and $p_t$ are related via a 2D homography: $p_t = h(p_0)$ where $h$ is taken as short notation for $h_{{\bf T},{\bf R},{\bf n}}(\cdot)$, to indicate that it is a function of ${\bf R}, {\bf T}, {\bf n}$ where ${\bf n}$ is the surface normal to the patch.

The aim is here for patches that are localized spatially, and whose texture is optimized to fool an image/scene classifier with assumed known architecture and weights, i.e., using a white box threat model (a best case scenario for the patch-generator). The patch attacks a network to force the prediction toward an incorrect and specific class label (targeted attack). Extending approaches in ~\cite{athalye2017synthesizing,brown2017adversarial},
a patch attack uses an optimized pattern that is texture-mapped onto a 3D co-planar patch which is applied to original image $I$ (using $\mathcal{A}_T(\cdot)$), and constrained to lie within a contiguous surface in 3D. This produces an adversarial exemplar image, $I'$, such that, when successful, the attacked classifier will produce a correct classification $\Phi(I) = \hat{y} = y$ for $I$, but an incorrect targeted label classification, $\Phi(I') = \hat{y} = y_t \neq y$.

The perturbation pattern in the patch $q$ is found by solving an optimization process expressed as an expectation over the desired combination of image classes and likely 3D geometric transformations:

\begin{equation}
\label{eq:patch_EoT}
    \argmax_{q} \mathbb{E}_{I\sim{{\cal I}}, {\bf R}\sim{{\cal R}}, {\bf T}\sim{{\cal T}}, d\sim{D}}~\Phi(y_t|\mathcal{A}_{{\bf R}, {\bf T}, {\bf n}}(q, I))
\end{equation}

where 
\begin{itemize}
    \item $\mathcal{A}_{{\bf R}, {\bf T}, {\bf n}}(q, I) = D(c(\mathcal{W}_{{\bf R}, {\bf T}, {\bf n}}(q)), I)$ is a patch applicator transformation that is the result of composing a set of functions such that:
        \begin{itemize}
            \item $\mathcal{W}_{ {\bf R}, {\bf T}, {\bf n}}(q)$ denotes the image domain texture mapping and then warping of a co-planar patch $q$ under a homography $h_{\bf T, R, n}$ that results from a rotation $R$, translation $T$, and 3D patch plane orientation ${\bf n}$.
            \item $c(\cdot)$ denotes the point projection operation modeling the camera projection (as in Equation~\ref{eq:point_proj}),
            \item $D(q,I)$ is a function that performs 2D digital insertion of a patch $q$ into the scene image $I$,
            \item ${\cal R} \times {\cal T}  \subset SE(3)$ are the subset of rotations and translations over which to optimize the patch and can, for example, correspond to the set of rotations and translations seen during the patch motion,
            \item ${\cal I}$ denotes the set of images over which patches will be inserted and optimized.
        \end{itemize}
\end{itemize}

In plain terms, and looking at Figure~\ref{fig:constrained_eot}, consider the case of a 3D patch attack aimed at fooling an autonomous driving system to be misled to classify stop signs as speed limits, then the optimization seeks to find the patch pattern that best leads to a misclassification $\hat{y}=y_{SL}$ (targeting the class \textit{speed limit signs (SL)}), and takes into consideration the expectation over a set of images ${\cal I}$ sampled over a set of image classes to be fooled (e.g., the classes including `stop sign' and `yield signs'). During optimization, the process warps the patch according to a set of homographies where the rotation and translation were sampled over transformations of interest such as those consistent with vehicle approach trajectories.

\subsection{Mean Attack Success over Transformations (mAST)}
In order to provide a holistic measure of the effectiveness of a patch attack over a given range of transformations, we define the following \textit{mean Attack Success over Transformations (mAST)} metric:

\begin{gather*}
\label{eq:patch_AUC}
    mAST_\phi = \frac{1}{|{\cal C}|} \sum\limits_{c\in \mathcal{C}} \frac{1}{\beta-\alpha} \int _\alpha ^\beta S_c(\phi) d\phi \\
    mAST_\phi \in [0, 1]
\end{gather*}

\noindent where $\phi \in [\alpha, \beta]$ is the value of the transformation parameter (e.g., roll angle) under consideration, and $S_c(\phi)$ refers to the success rate of the given patch with target label $c$ (from the set of targets ${\cal C}$) when tested at transformation $\phi$. 

This $mAST_\phi$ metric provides an aggregate score for patch success when considering both varying target labels (via ${\cal C}$) and a set of optimization strategies where the train time support may differ (via varying $\phi$).

\section{Experiments}
\label{sec:EoT_experiments}

\subsection{Experimental Protocol}
To test the robustness of patch attacks over a range of 3D poses, we use the following experimental protocol for evaluation.  Similar to~\cite{jackofalltrades}, we assess the patch attack effectiveness over a range of training/testing transformations and in particular, focus on quantifying the impact of differences between the training and testing transformation distributions.  We run the optimization/evaluation in a white-box setting against a ResNet-50~\cite{he2016deep} model trained on ImageNet, which represents a best case scenario from the patch attack scenario (i.e., where full knowledge of architecture and weights is available to the attacker). 

Following~\cite{jackofalltrades}, we first select a set of transformations of interest (i.e., yaw, roll, loom/scale) and define the train time support $\mathcal{T} \sim \mathcal{U}_{\psi, \theta, z}$ (with a slight abuse of notation for $\mathcal{T}$) where the set of transformations is drawn uniformly random from pre-defined ranges (e.g., $yaw \triangleq \psi \in [\pm20\degree]$; $roll \triangleq \theta \in [\pm10\degree]$; $loom \triangleq z \in [6,8]$). Then, we optimize patch attacks (following~\cite{brown2017adversarial}), specifically targeting 15 ImageNet classes. Through initial experimentation, we identified subsets of ImageNet classes which tend to achieve higher/lower success rates when used as patch targets and selected five targeted classes from three tiers defined by high, medium, and low effectiveness.

All patches are trained for a single epoch consisting of 200 batches of 32 images from the ILSVRC 2012 validation set. After optimization, each attack is then tested against a set of transformations covering the full range of motion (e.g., $\psi \in [\pm180\degree]$), with each test transformation applied to patches inserted into 320 randomly sampled images from the ImageNet test set. Patch coordinates are randomized during optimization and at attack time unless specified otherwise.
The field of view (FoV) of the camera is kept constant and the camera centered at (0,0). FoV is set to $60 \degree$.

In this approach, we consider differences between train and test-time transformation distributions so as to estimate whether this disparity improves or degrades patch success.  We organize our experiments so as to answer a set of important questions meant to elucidate core behavior of patch attacks under independent (e.g., yaw alone) and composed sets of transformations (e.g., yaw and roll together).

\begin{figure*}[ht]
    \includegraphics[scale=0.39]{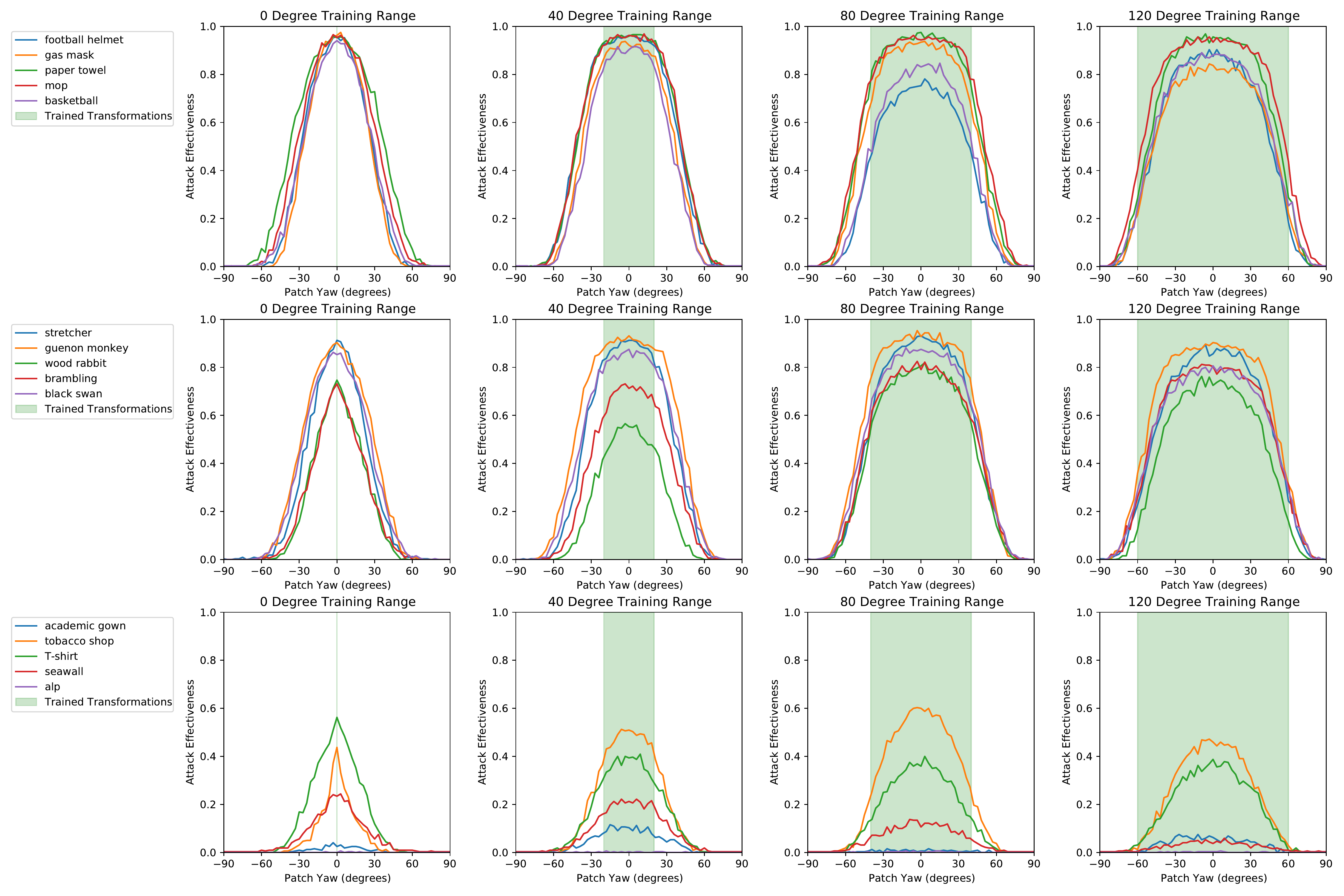}
    \caption{Performance results obtained with patches optimized over a range of yaw angles (shown in the green shaded area). Attack effectiveness is shown on the Y axes as a function of actual yaw angle testing was performed at (X axes). Top, middle and bottom rows present the aggregation of results grouping together the high-, mid-, and low- performing target classes respectively.}
    \label{fig:yaw_acc}
\end{figure*}

\begin{figure*}[ht]
    \includegraphics[scale=0.39]{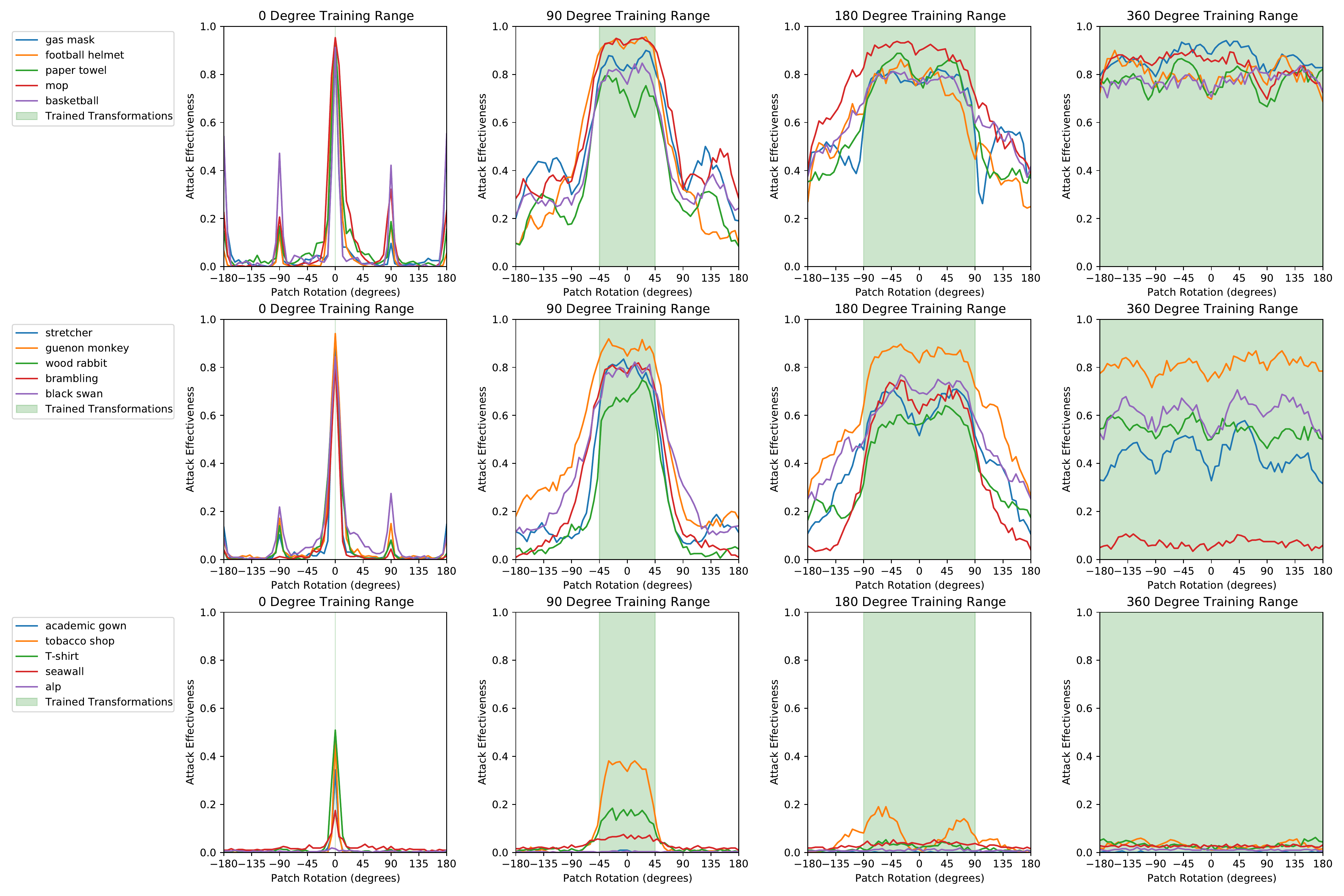}
    \caption{Performance results obtained with patches optimized over a range of in-plane rotations (shown in the green shaded area). Similarly to Figure 3, attack effectiveness is on the Y axes as a function of rotation angle (X axes). Top, middle and bottom rows present the aggregation of results grouping together the high-, mid-, and low- performing target classes respectively.}
    \label{fig:roll_acc}
\end{figure*}

\begin{figure*}[ht]
    \includegraphics[scale=0.38]{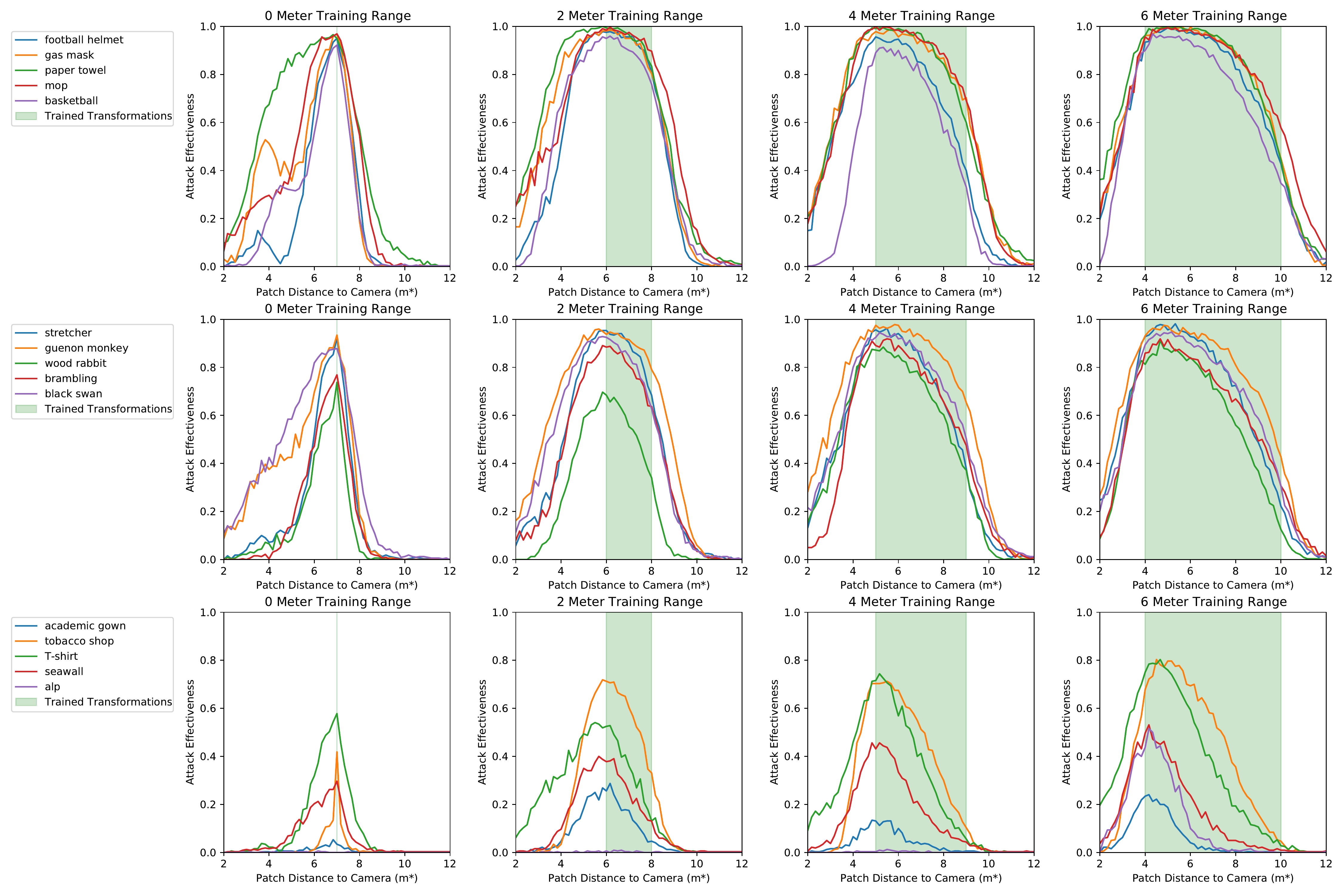}
    \caption{Performance results obtained with patches optimized over a range of simulated camera distances (shown in the green shaded area). Attack effectiveness is on the Y axes as a function of camera distance (X axes). Top, middle and bottom rows present the aggregation of results for the high-, mid-, and low- performing target classes respectively. Note: although the distances here are given in meters for ease of labeling, the units are arbitrary and solely meant to allow comparison of different distances within the experiment.}
    \label{fig:loom_acc}
\end{figure*}

\subsection{Influence of Yaw Motion}
\label{sub:yaw_exp}

{\bf Question:} As a targeted vision system (such as on a self-driving car) moves with respect to a stationary patch, perceived out-of-plane rotations of the patch will occur if the direction of motion is not normal to the surface of the patch. Thus, we ask: {\it what is the impact of out-of-plane transformations (pitch and yaw) on patch success? How does optimizing over a more aggressive range of Euler angles impact patch effectiveness and/or robustness to such transformations?}

{\bf Approach:} These questions address whether the patch optimization process produces patterns that rely on precise orientation relative to the camera or can withstand some degree of foreshortening. From this point forward when discussing out-of-plane rotations, we only refer to yaw, and not pitch since they are essentially equivalent in this context. We argue that the choice of rotation axis (in this case, $x$ vs. $y$, where the camera is looking along the $z$-axis) does not impact the optimization process, and that we can focus on one type of out-of-plane rotation without loss of generality. To address the yaw questions, we optimize a set of patches following the experimental protocol of Section~\ref{sec:EoT_experiments}, with patch roll fixed at $0 \degree$, location randomized, and camera distance set to 7 meters. We perform experiments for four levels of optimization with yaw transformations; namely, we optimize patches with train time support, $\mathcal{T} \sim \mathcal{U}_{\psi_i, \theta, z}$ where $\psi_i \in \{[\pm0\degree], [\pm20\degree], [\pm40\degree], [\pm60\degree] \}$, $\theta \in [\pm0\degree]$, and $z \in [7, 7]$ (where $\mathcal{U}_{\psi_i, \theta, z}$ denotes a multivariate uniform distribution sampled according to the intervals specified by $\psi_i, \theta, z$). All experiments were evaluated over a yaw range of $180\degree$ subdivided into 60 intervals of $3\degree$.

{\bf Experiments:} Results for the yaw experiments are captured in Figure~\ref{fig:yaw_acc}.
\begin{table}[h]
    \centering
    \resizebox{\columnwidth}{!}{ 
    \begin{tabular}{|c|c|c|c|c|} 
    \hline
        Yaw $(\psi)$ Training Range & $\pm 0\degree$ & $\pm 20\degree$ & $\pm 40\degree$ & $\pm 60\degree$ \\ \hline
        Top 5 Avg & 0.34 & 0.43 & 0.48 & 0.52 \\ \hline
        Middle 5 Avg & 0.26 & 0.34 & 0.45 & 0.47 \\ \hline
        Bottom 5 Avg & 0.06 & 0.09 & 0.09 & 0.07 \\ \hline
    \end{tabular}
    }
    \caption{$mAST_\psi$ for Experiment~\ref{sub:yaw_exp}}
    \label{tab:yaw_exp}
\end{table}

{\bf Discussion:} From this experiment, we observe a consistently steep drop in patch effectiveness once the magnitude of the patch yaw reaches approximately $60\degree$. This appears to be the consequence of severe foreshortening having a cutoff effect on the patch ability to attack.  This effect is in sharp contrast to past sensitivity analysis based on 2D transformation as reported in~\cite{jackofalltrades}. While optimizing over wider transformation support appears to extend the interval on which patch success reaches at least $90\%$, it is clear that by $\pm 70\degree$ the patch success drops to zero in almost every case. This is presumably due to the high degree of foreshortening experienced by a planar patch rotated $60 \degree$ or more away from the camera; see Figure 2 for an example of a patch rotated at yaw angle $60\degree$.

\subsection{Influence of Roll Motion}
\label{sub:roll_exp}
{\bf Question:} In a physical attack scenario (e.g., vehicular motion), changes to the environment or to the surface on which the patch is placed may result in changes in the patch's orientation (e.g. road surface pot-holes yielding roll motion due to the car suspension system). Thus, we ask: {\it what is the impact of in-plane rotations (roll) on patch success? How does greater freedom of rotation at train time impact patch effectiveness and/or robustness to such transformations?}

{\bf Approach:} To address the roll questions, we again follow the base experiment setup, with patch yaw fixed at $0 \degree$, location randomized, and camera distance set to 7 meters. Taking the angle of rotation of the patch to be $\theta$, we train patches with train time support $\mathcal{T} \sim \mathcal{U}_{\psi, \theta_i, z}$ where $\theta_i \in \{ [\pm0\degree], [\pm45\degree], [\pm90\degree], [\pm180\degree] \}$, $\psi \in [\pm0\degree]$, and $z \in [7,7]$. All experiments were evaluated over a roll range of $360 \degree$ subdivided into 60 intervals.

{\bf Experiments:} Results for the roll experiments are captured in Figure~\ref{fig:roll_acc}.

\begin{table}[h]
    \centering
    \resizebox{\columnwidth}{!}{ 
    \begin{tabular}{|c|c|c|c|c|} 
    \hline
        Roll $(\theta)$ Training Range & $\pm 0\degree$ & $\pm 45\degree$ & $\pm 90\degree$ & $\pm 180\degree$ \\ \hline
        Top 5 Avg & 0.08 & 0.49 & 0.65 & 0.81 \\ \hline
        Middle 5 Avg & 0.06 & 0.35 & 0.50 & 0.49 \\ \hline
        Bottom 5 Avg & 0.01 & 0.04 & 0.02 & 0.02 \\ \hline
    \end{tabular}
    }
    \caption{$mAST_\theta$ for Experiment~\ref{sub:roll_exp}}
    \label{tab:roll_exp}
\end{table}

{\bf Discussion:} In these results, a different pattern emerges as the result of patch training. With no rotation present in the patch training, the performance approaches $100\%$ at $0\degree$ in the high- and mid-performing patches, but falls off immediately when any rotation is introduced at test time. When rotations are permitted in training, the performance across the trained range improves overall, but the peak is lowered. This pattern is especially evident in the mid-tier patches, where the patches trained on no rotation peak sharply at $0\degree$, while the patches trained on $360\degree$ of rotation perform consistently in the 0.4 - 0.8 range across all rotations.

\subsection{Influence of Looming Motion}
\label{sub:loom_exp}
{\bf Question:} In a real world attack, the distance (and thus apparent scale) of the patch is likely to vary, and certainly in the case of an autonomous vehicle vision system which is moving toward the patch. Thus, we ask: {\it what is the impact of camera distance (loom) on patch success? How does training on different camera distances impact patch effectiveness and/or robustness to such changes in apparent scale?}

{\bf Approach:} To address the loom questions, we follow the base experiment setup, with patch yaw fixed at $0 \degree$, patch roll fixed at $0\degree$, and location randomized. Here, we vary the position of the patch (given in meters, although this is merely a convenience unit) along the z-axis. We train patches with train time support $\mathcal{T} \sim \mathcal{U}_{\psi, \theta, z_i}$ where $\psi \in [\pm0\degree]$, $\theta \in [\pm0\degree]$, and $z_i \in \{[6,8], [5,9], [4,10]\}$. All experiments were evaluated over a 10 meter loom range subdivided into $60$ intervals.

{\bf Experiments:} Results for the loom experiments are captured in Figure~\ref{fig:loom_acc}.

\begin{table}[h]
    \centering
    \resizebox{\columnwidth}{!}{ 
    \begin{tabular}{|c|c|c|c|c|}
    \hline
        Loom $(z)$ Training Range& [7, 7] & [6, 8] & [5, 9] & [4, 10] \\ \hline
        Top 5 Avg & 0.31 & 0.53 & 0.56 & 0.67 \\ \hline
        Middle 5 Avg & 0.22 & 0.41 & 0.52 & 0.57 \\ \hline
        Bottom 5 Avg & 0.04 & 0.13 & 0.14 & 0.20 \\ \hline
    \end{tabular}
    }
    \caption{$mAST_z$ for Experiment~\ref{sub:loom_exp}}
    \label{tab:loom_exp}
\end{table}

{\bf Discussion:} From these results, it is clear that increasing the range of camera distances seen during training expands the region of scales at which the patch is effective at test time. However, another effect also emerges, wherein larger patch scales outperform smaller ones, even within the bounds of the same training interval. This is logical, given the higher possible level of detail of patch features with a larger patch, as well as the increased prevalence within the image and higher chance of occlusion of the base image class.

\subsection{Combination of Roll and Yaw Motion}
\label{sub:combo}
{\bf Question:} In a physical attack scenario, changes to the patch's orientation in 3D space may not occur in isolation, but instead result in apparent combinations of multiple transformations. Thus, we ask: {\it how do in-plane rotations (roll) and out-of-plane rotations (yaw) combine to impact patch success? How does a greater range of motion in both rotations at train time impact patch effectiveness and/or robustness to such compound transformations?}

{\bf Approach:} To address the compound questions, we again follow the base experiment setup, with location randomized and camera distance set to 7 meters. Taking the yaw angle of the patch to be $\psi$, and the roll angle to be $\theta$, we train patches with train time support $\mathcal{T} \sim \mathcal{U}_{\psi_i, \theta_i, z}$ where $\psi_i \in \{[\pm0\degree], [\pm20\degree], [\pm40\degree], [\pm60\degree] \}$, $\theta_i \in \{[\pm0\degree], [\pm45\degree], [\pm90\degree], [\pm180\degree] \}$, and $z \in [7,7]$. All experiments were evaluated over a grid of transformations,  covering a yaw range of $\pm180 \degree$ and a roll range of $\pm360 \degree$, subdivided into 400 intervals ($20 \times 20$, so $21 \times 21$ endpoints).

\begin{figure*}[ht]
    \includegraphics[scale=0.38]{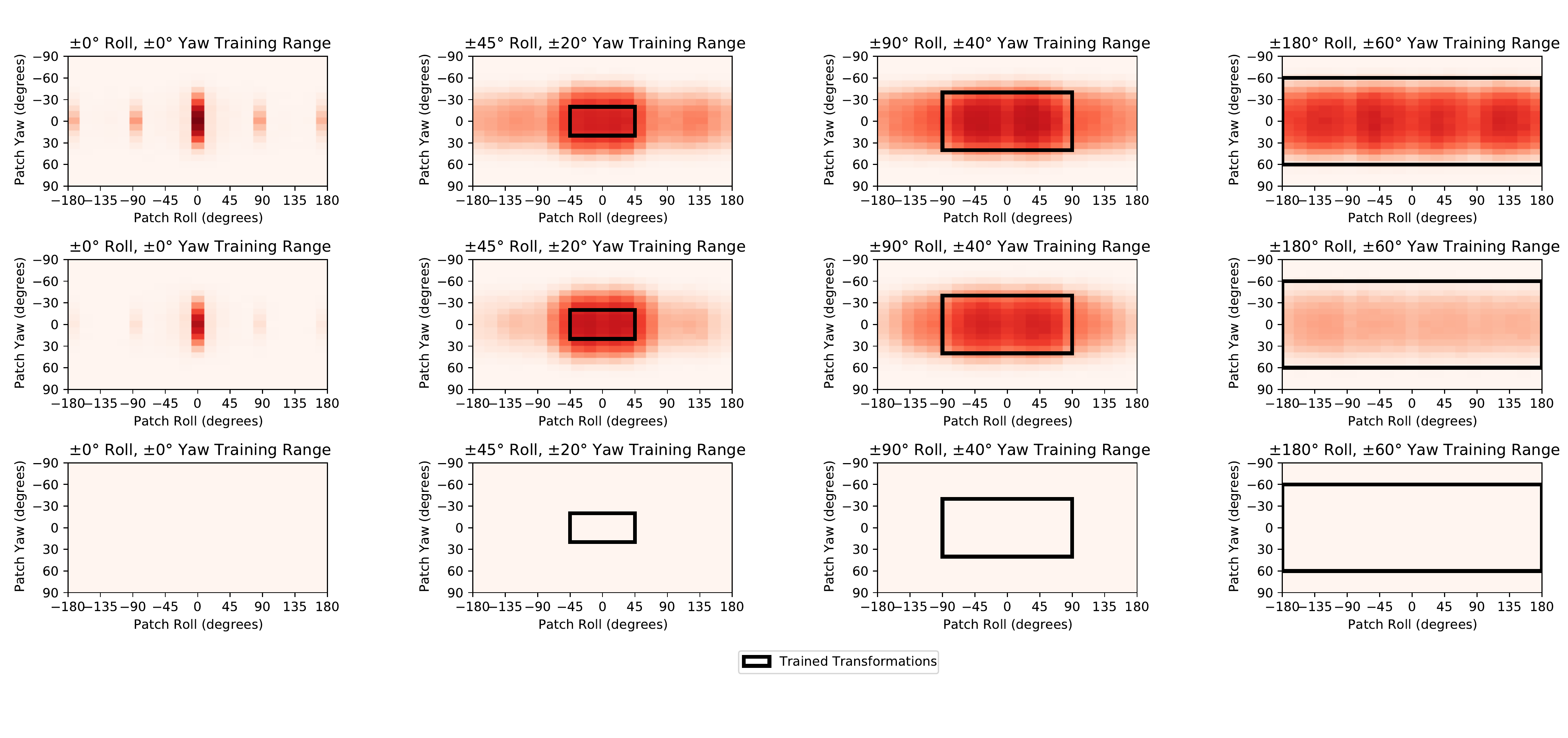}
    \caption{Performance results obtained with patches optimized over a combination of in-plane and out-of-plane rotations (shown via a box overlaid on each plot). Since there are two variables under investigation, performance is conveyed via a heatmap, with colors representing a scale from 0 (white) to 1 (red). Top, middle and bottom rows present the aggregation of results grouped together according to the high-, mid-, and low- performing target classes.}
    \label{fig:compound_acc}
\end{figure*}

{\bf Experiments:} Results for the roll experiments are captured in Figure~\ref{fig:compound_acc}.

{\bf Discussion:} Despite the compositional nature of the transformations, the patterns of patch behavior along each individual axis closely mirror the results obtained from analyzing that transformation separately, indicating minimal interactions between the pair of transformations. This is particularly evident for roll, since the multimodal pattern of behavior is consistent with the behavior in Figure~\ref{fig:roll_acc}. More specifically, peaks are not only present but peak accuracy values occur at the same angles for each respective cell. Looking to the vertical axis, the regions of peak effectiveness become increasingly spread out when examining the plots from left to right along a given row.  This result parallels those of Figure~\ref{fig:yaw_acc}, indicating a clear improvement from increasing the yaw training distribution.

\section{Discussion}
The results of experiments in Section~\ref{sec:EoT_experiments} demonstrate several overall trends. In all experiments, extending the train time support leads to measurable gains in patch success over all test time transformations.  While yaw shows marginal improvement beyond a window of $\pm30\degree$ due to foreshortening effects, roll and loom exhibit much stronger gains as the train time support increases. Our novel $mAST_\phi$ metric provides a means to accurately quantify and confirm these trends as seen in Tables~\ref{tab:yaw_exp},~\ref{tab:roll_exp}, and~\ref{tab:loom_exp}.  

Due to the targeted nature of patch attacks, these experiments highlight the importance of choosing a suitable target class label to achieve high attack success. In analyzing the effects of 3D transformations on patch success, we build upon prior literature and further confirm this sensitivity.  Results such as in Figure~\ref{fig:roll_acc} illustrate increasing separation in patch success across a range of target labels. Further study around identifying maximally-attacking target labels remains important and we leave it to future work to further study this phenomenon.

Additionally, experiments in Section~\ref{sub:combo} suggest few interactions between in- and out-of-plane rotations with respect to patch success. While a valuable insight in its own right, this observation underscores the importance of performing a principled evaluation of individual transformations as in the preceding sections.  Without those prior experiments, the outcomes of Section~\ref{sub:combo} would not allow any clear conclusions about whether behavior of composed transformations deviates significantly from their isolated counterparts.

Lastly, we recognize that our experiments are not without limitations. Our experiments are certainly not comprehensive, and we leave it to future work to improve the patch optimization and evaluation process to further study assess the effects of modeling lighting, color variation, scene geometry, and other aspects of achieving patch/scene realism. Additionally, as our method considers only average-case performance of the patch attack, we leave it to future work to consider worst-case performance with respect to the patch and sampled transformations in performing the optimization.

\section{Conclusions}
This paper examines the incidence of 3D pose on the attackness of adversarial patches and introduces a new metric characterizing patch attack robustness. This study provides an analysis of the invariance of patch attacks for 3D real world applications, which had been largely missing in the literature. It illustrates that 3D robustness to pose is highly dependent on targeted class label, and also demonstrates, in contrast to~\cite{jackofalltrades}, that optimizing over larger range generally provides benefits. It also shows that there is a fundamental limit to this behavior with a cutoff in improvement beyond certain out of plane angles. Lastly, the analysis of the effect of composed transformations on patch attackness indicated few, if any, interactions between transformations, an insight which could only have been gained through the principled evaluation of individual transformations. These behaviors should help inform future work in the design of 3D patch attacks and defenses.

\begin{figure*}[h]
\includegraphics[scale=0.67]{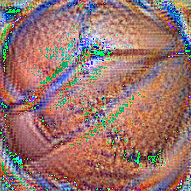}
\includegraphics[scale=0.67]{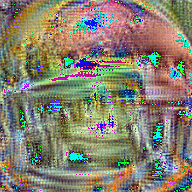}
\includegraphics[scale=0.67]{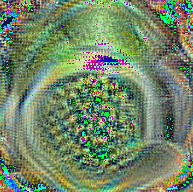}
\includegraphics[scale=0.67]{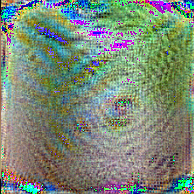}
\includegraphics[scale=0.67]{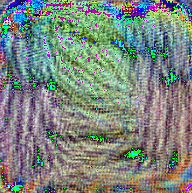}
\caption{Examples of 5 of the top 15 best-performing patch target classes, according to our analysis.}

\includegraphics[scale=0.67]{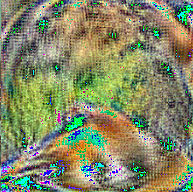}
\includegraphics[scale=0.67]{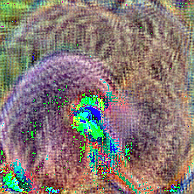}
\includegraphics[scale=0.67]{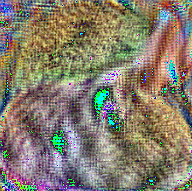}
\includegraphics[scale=0.67]{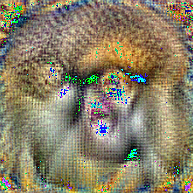}
\includegraphics[scale=0.67]{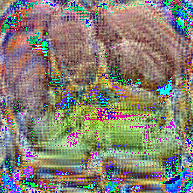}
\caption{Examples of 5 of the 15 patch classes centered around the mean effectiveness.}

\includegraphics[scale=0.67]{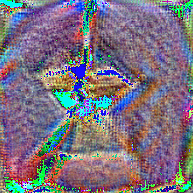}
\includegraphics[scale=0.67]{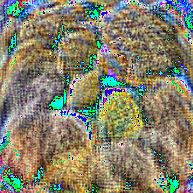}
\includegraphics[scale=0.67]{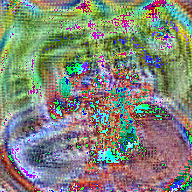}
\includegraphics[scale=0.67]{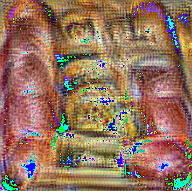}
\includegraphics[scale=0.67]{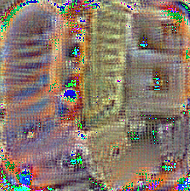}
\caption{Examples of 5 of the 15 least effective patch classes.}
\end{figure*}

{\small
\bibliographystyle{ieee_fullname}
\bibliography{egbib}

\begin{thebibliography}{10}\itemsep=-1pt

\bibitem{athalye2017synthesizing}
Anish Athalye, Logan Engstrom, Andrew Ilyas, and Kevin Kwok.
\newblock Synthesizing robust adversarial examples.
\newblock {\em arXiv preprint arXiv:1707.07397}, 2017.

\bibitem{brown2017adversarial}
Tom~B Brown, Dandelion Man{\'e}, Aurko Roy, Mart{\'\i}n Abadi, and Justin
  Gilmer.
\newblock Adversarial patch.
\newblock {\em arXiv preprint arXiv:1712.09665}, 2017.

\bibitem{burlina1995context}
P Burlina, R Chellappa, C Lin, and X Zhang.
\newblock Context-based exploitation of aerial imagery.
\newblock In {\em Proc. Workshop on Model-Based Vision (Boston, MA)}, 1995.

\bibitem{burlina2018utility}
Phillippe Burlina, Neil Joshi, Katia~D Pacheco, David~E Freund, Jun Kong, and
  Neil~M Bressler.
\newblock Utility of deep learning methods for referability classification of
  age-related macular degeneration.
\newblock {\em JAMA ophthalmology}, 136(11):1305--1307, 2018.

\bibitem{burlina2020addressing}
Philippe Burlina, Neil Joshi, William Paul, Katia~D Pacheco, and Neil~M
  Bressler.
\newblock Addressing artificial intelligence bias in retinal disease
  diagnostics.
\newblock {\em arXiv preprint arXiv:2004.13515}, 2020.

\bibitem{burlina2020low}
Philippe Burlina, William Paul, Philip Mathew, Neil Joshi, Katia~D Pacheco, and
  Neil~M Bressler.
\newblock Low-shot deep learning of diabetic retinopathy with potential
  applications to address artificial intelligence bias in retinal diagnostics
  and rare ophthalmic diseases.
\newblock {\em JAMA ophthalmology}, 138(10):1070--1077, 2020.

\bibitem{burlina2019automated}
Philippe~M Burlina, Neil~J Joshi, Elise Ng, Seth~D Billings, Alison~W Rebman,
  and John~N Aucott.
\newblock Automated detection of erythema migrans and other confounding skin
  lesions via deep learning.
\newblock {\em Computers in biology and medicine}, 105:151--156, 2019.

\bibitem{carlini2017adversarial}
Nicholas Carlini and David Wagner.
\newblock Adversarial examples are not easily detected: Bypassing ten detection
  methods.
\newblock In {\em Proceedings of the 10th ACM Workshop on Artificial
  Intelligence and Security}, pages 3--14, 2017.

\bibitem{chen2018shapeshifter}
Shang-Tse Chen, Cory Cornelius, Jason Martin, and Duen Horng~Polo Chau.
\newblock Shapeshifter: Robust physical adversarial attack on faster r-cnn
  object detector.
\newblock In {\em Joint European Conference on Machine Learning and Knowledge
  Discovery in Databases}, pages 52--68. Springer, 2018.

\bibitem{czaja2018adversarial}
Wojciech Czaja, Neil Fendley, Michael Pekala, Christopher Ratto, and I-Jeng
  Wang.
\newblock Adversarial examples in remote sensing.
\newblock In {\em Proceedings of the 26th ACM SIGSPATIAL International
  Conference on Advances in Geographic Information Systems}, pages 408--411,
  2018.

\bibitem{eykholt2018robust}
Kevin Eykholt, Ivan Evtimov, Earlence Fernandes, Bo Li, Amir Rahmati, Chaowei
  Xiao, Atul Prakash, Tadayoshi Kohno, and Dawn Song.
\newblock Robust physical-world attacks on deep learning visual classification.
\newblock In {\em Proceedings of the IEEE conference on computer vision and
  pattern recognition}, pages 1625--1634, 2018.

\bibitem{jackofalltrades}
Neil Fendley, Max Lennon, I-Jeng Wang, Philippe Burlina, and Nathan Drenkow.
\newblock Jacks of all trades, masters of none: Addressing distributional shift
  and obtrusiveness via transparent patch attacks.
\newblock 2019.

\bibitem{geng2020recent}
Chuanxing Geng, Sheng-jun Huang, and Songcan Chen.
\newblock Recent advances in open set recognition: A survey.
\newblock {\em IEEE Transactions on Pattern Analysis and Machine Intelligence},
  2020.

\bibitem{goodfellow2014explaining}
Ian~J Goodfellow, Jonathon Shlens, and Christian Szegedy.
\newblock Explaining and harnessing adversarial examples.
\newblock {\em arXiv preprint arXiv:1412.6572}, 2014.

\bibitem{he2016deep}
Kaiming He, Xiangyu Zhang, Shaoqing Ren, and Jian Sun.
\newblock Deep residual learning for image recognition.
\newblock In {\em Proceedings of the IEEE conference on computer vision and
  pattern recognition}, pages 770--778, 2016.

\bibitem{krizhevsky2017imagenet}
Alex Krizhevsky, Ilya Sutskever, and Geoffrey~E Hinton.
\newblock Imagenet classification with deep convolutional neural networks.
\newblock {\em Communications of the ACM}, 60(6):84--90, 2017.

\bibitem{kurakin2016adversarial}
Alexey Kurakin, Ian Goodfellow, Samy Bengio, et~al.
\newblock Adversarial examples in the physical world, 2016.

\bibitem{lecun2015deep}
Yann LeCun, Yoshua Bengio, and Geoffrey Hinton.
\newblock Deep learning.
\newblock {\em nature}, 521(7553):436, 2015.

\bibitem{mnih2015human}
Volodymyr Mnih, Koray Kavukcuoglu, David Silver, Andrei~A Rusu, Joel Veness,
  Marc~G Bellemare, Alex Graves, Martin Riedmiller, Andreas~K Fidjeland, Georg
  Ostrovski, et~al.
\newblock Human-level control through deep reinforcement learning.
\newblock {\em Nature}, 518(7540):529, 2015.

\bibitem{moosavi2017universal}
Seyed-Mohsen Moosavi-Dezfooli, Alhussein Fawzi, Omar Fawzi, and Pascal
  Frossard.
\newblock Universal adversarial perturbations.
\newblock In {\em Proceedings of the IEEE conference on computer vision and
  pattern recognition}, pages 1765--1773, 2017.

\bibitem{moosavi2016deepfool}
Seyed-Mohsen Moosavi-Dezfooli, Alhussein Fawzi, and Pascal Frossard.
\newblock Deepfool: a simple and accurate method to fool deep neural networks.
\newblock In {\em Proceedings of the IEEE conference on computer vision and
  pattern recognition}, pages 2574--2582, 2016.

\bibitem{pekala2019deep}
Mike Pekala, Neil Joshi, TY~Alvin Liu, Neil~M Bressler, D~Cabrera DeBuc, and
  Philippe Burlina.
\newblock Deep learning based retinal oct segmentation.
\newblock {\em Computers in Biology and Medicine}, 114:103445, 2019.

\bibitem{ravi2016optimization}
Sachin Ravi and Hugo Larochelle.
\newblock Optimization as a model for few-shot learning.
\newblock 2016.

\bibitem{Redmon2015YouOL}
Joseph Redmon, Santosh~Kumar Divvala, Ross~B. Girshick, and Ali Farhadi.
\newblock You only look once: Unified, real-time object detection.
\newblock {\em 2016 IEEE Conference on Computer Vision and Pattern Recognition
  (CVPR)}, pages 779--788, 2015.

\bibitem{Ren2015FasterRT}
Shaoqing Ren, Kaiming He, Ross~B. Girshick, and Jian Sun.
\newblock Faster r-cnn: Towards real-time object detection with region proposal
  networks.
\newblock {\em IEEE Transactions on Pattern Analysis and Machine Intelligence},
  39:1137--1149, 2015.

\bibitem{ronneberger2015u}
Olaf Ronneberger, Philipp Fischer, and Thomas Brox.
\newblock U-net: Convolutional networks for biomedical image segmentation.
\newblock In {\em International Conference on Medical image computing and
  computer-assisted intervention}, pages 234--241. Springer, 2015.

\bibitem{scheirer2012toward}
Walter~J Scheirer, Anderson de Rezende~Rocha, Archana Sapkota, and Terrance~E
  Boult.
\newblock Toward open set recognition.
\newblock {\em IEEE transactions on pattern analysis and machine intelligence},
  35(7):1757--1772, 2012.

\bibitem{shah2018susceptibility}
Abhay Shah, Stephanie Lynch, Meindert Niemeijer, Ryan Amelon, Warren Clarida,
  James Folk, Stephen Russell, Xiaodong Wu, and Michael~D Abr{\`a}moff.
\newblock Susceptibility to misdiagnosis of adversarial images by deep learning
  based retinal image analysis algorithms.
\newblock In {\em 2018 IEEE 15th International Symposium on Biomedical Imaging
  (ISBI 2018)}, pages 1454--1457. IEEE, 2018.

\bibitem{sharif2016accessorize}
Mahmood Sharif, Sruti Bhagavatula, Lujo Bauer, and Michael~K Reiter.
\newblock Accessorize to a crime: Real and stealthy attacks on state-of-the-art
  face recognition.
\newblock In {\em Proceedings of the 2016 acm sigsac conference on computer and
  communications security}, pages 1528--1540, 2016.

\bibitem{shokri2017membership}
Reza Shokri, Marco Stronati, Congzheng Song, and Vitaly Shmatikov.
\newblock Membership inference attacks against machine learning models.
\newblock In {\em 2017 IEEE Symposium on Security and Privacy (SP)}, pages
  3--18. IEEE, 2017.

\bibitem{silver2017mastering}
David Silver, Julian Schrittwieser, Karen Simonyan, Ioannis Antonoglou, Aja
  Huang, Arthur Guez, Thomas Hubert, Lucas Baker, Matthew Lai, Adrian Bolton,
  et~al.
\newblock Mastering the game of go without human knowledge.
\newblock {\em nature}, 550(7676):354--359, 2017.

\bibitem{szegedy2013intriguing}
Christian Szegedy, Wojciech Zaremba, Ilya Sutskever, Joan Bruna, Dumitru Erhan,
  Ian Goodfellow, and Rob Fergus.
\newblock Intriguing properties of neural networks.
\newblock {\em arXiv preprint arXiv:1312.6199}, 2013.

\bibitem{thys2019fooling}
Simen Thys, Wiebe Van~Ranst, and Toon Goedem{\'e}.
\newblock Fooling automated surveillance cameras: adversarial patches to attack
  person detection.
\newblock In {\em Proceedings of the IEEE/CVF Conference on Computer Vision and
  Pattern Recognition Workshops}, pages 0--0, 2019.

\bibitem{vyas2013estimating}
Saurabh Vyas, Amit Banerjee, and Philippe Burlina.
\newblock Estimating physiological skin parameters from hyperspectral
  signatures.
\newblock {\em Journal of biomedical optics}, 18(5):057008, 2013.

\bibitem{wu2020dpattack}
Shudeng Wu, Tao Dai, and Shu-Tao Xia.
\newblock Dpattack: Diffused patch attacks against universal object detection.
\newblock {\em arXiv preprint arXiv:2010.11679}, 2020.

\bibitem{wu2020physical}
Tong Wu, Xuefei Ning, Wenshuo Li, Ranran Huang, Huazhong Yang, and Yu Wang.
\newblock Physical adversarial attack on vehicle detector in the carla
  simulator.
\newblock {\em arXiv preprint arXiv:2007.16118}, 2020.

\bibitem{wu2020making}
Zuxuan Wu, Ser-Nam Lim, Larry~S Davis, and Tom Goldstein.
\newblock Making an invisibility cloak: Real world adversarial attacks on
  object detectors.
\newblock In {\em European Conference on Computer Vision}, pages 1--17.
  Springer, 2020.

\bibitem{zhao2020object}
Yusheng Zhao, Huanqian Yan, and Xingxing Wei.
\newblock Object hider: Adversarial patch attack against object detectors.
\newblock {\em arXiv preprint arXiv:2010.14974}, 2020.

\bibitem{zolfi2021translucent}
Alon Zolfi, Moshe Kravchik, Yuval Elovici, and Asaf Shabtai.
\newblock The translucent patch: A physical and universal attack on object
  detectors.
\newblock In {\em Proceedings of the IEEE/CVF Conference on Computer Vision and
  Pattern Recognition}, pages 15232--15241, 2021.

\end{thebibliography}
}
\end{document}